\documentclass[11pt,letterpaper]{article}
\usepackage{emnlp2016}
\usepackage{times}
\usepackage{latexsym}

\usepackage{tabularx}
\usepackage{caption}
\usepackage{graphicx}
\usepackage{bm}
\usepackage{algorithm}
\usepackage{algorithmic}
\usepackage{subfigure}
\usepackage{bbm}
\usepackage{multirow}
\usepackage{epstopdf}

\usepackage{color}
\newcommand*{\affaddr}[1]{#1}
\newcommand*{\affmark}[1][*]{\textsuperscript{#1}}

\def\R{{\mathbbm R}}
\def\K{{\mathcal K}}

\emnlpfinalcopy

\title{Recursive Neural Conditional Random Fields \\for Aspect-based Sentiment Analysis}
\author{Wenya Wang\affmark[\dag\ddag]
        \hspace{3mm}
        Sinno Jialin Pan\affmark[\dag]
        \hspace{3mm}
        Daniel Dahlmeier\affmark[\ddag]
        \hspace{3mm}
        Xiaokui Xiao\affmark[\dag]
        \\
        \affaddr{\affmark[\dag]Nanyang Technological University, Singapore} \\
        \affaddr{\affmark[\ddag]SAP Innovation Center Singapore} \\
        \affmark[\dag]\{wa0001ya, sinnopan, xkxiao\}@ntu.edu.sg, \affmark[\ddag]\{d.dahlmeier\}@sap.com
        }
\date{}

\begin{document}

\maketitle

\begin{abstract}
In aspect-based sentiment analysis, extracting aspect terms along with the opinions being expressed from user-generated content is one of the most important subtasks. Previous studies have shown that exploiting connections between aspect and opinion terms is promising for this task. In this paper, we propose a novel joint model that integrates recursive neural networks and conditional random fields into a unified framework for explicit aspect and opinion terms co-extraction. The proposed model learns high-level discriminative features and double propagates information between aspect and opinion terms, simultaneously. Moreover, it is flexible to incorporate hand-crafted features into the proposed model to further boost its information extraction performance. Experimental results on the dataset from SemEval Challenge 2014 task 4 show the superiority of our proposed model over several baseline methods as well as the winning systems of the challenge.
\end{abstract}

\section{Introduction}\label{sec:intro}

Aspect-based sentiment analysis~\cite{Pang:2008:OMS:1454711.1454712,DBLP:series/dcsa/Liu11} aims to extract important information, e.g. opinion targets, opinion expressions, target categories, and opinion polarities, from user-generated content, such as microblogs, reviews, etc. This task was first studied by Hu and Liu~\shortcite{Hu04,Hu04b}, followed by~\cite{Pop05,Zhuang06,Zhang10,Qiu11,Li10}. In aspect-based sentiment analysis, one of the goals is to extract explicit aspects of an entity from text, along with the opinions being expressed. For example, in a restaurant review \textit{``I have to say they have one of the fastest delivery times in the city.''}, the aspect term is \textit{delivery times}, and the opinion term is \textit{fastest}.

Among previous work, one of the approaches is to accumulate aspect and opinion terms from a seed collection without label information, by utilizing syntactic rules or modification relations between them~\cite{Qiu11,DBLP:conf/webi/LiuGLZ13}. In the above example, if we know \textit{fastest} is an opinion word, then \textit{delivery times} is probably deduced as an aspect because \textit{fastest} is its modifier. However, this approach largely relies on hand-coded rules, and is restricted to certain Part-of-Speech (POS) tags, e.g., opinion words are restricted to be adjectives. Another approach focuses on feature engineering based on predefined lexicons, syntactic analysis, etc~\cite{Jin09,Li10}. A sequence labeling classifier is then built to extract aspect and opinion terms. This approach requires extensive efforts for designing hand-crafted features, and only combines features linearly for classification, which ignores higher order interactions.

To overcome the limitations of existing methods, we propose a novel model, namely Recursive Neural Conditional Random Fields (RNCRF). Specifically, RNCRF consists of two main components. The first component is to construct a recursive neural network (RNN)\footnote{Note that in this paper, RNN stands for recursive neural network instead of recurrent neural network.}~\cite{socher10} based on a dependency tree of each sentence. The goal is to learn a high-level feature representation for each word in the context of each sentence, and make the representation learning for aspect and opinion terms interactive through the underlying dependency structure among them. The output of the RNN is then fed into a Conditional Random Field (CRF)~\cite{Lafferty01} to learn a discriminative mapping from high-level features to labels, i.e., \textit{aspects}, \textit{opinions}, or \textit{others}, so that context information can be well captured. Our main contributions are to use RNN for encoding aspect-opinion relations in high-level representation learning, and to present a joint optimization approach based on maximum likelihood and backpropagation to learn the RNN and CRF components, simultaneously. In this way, the label information of aspect and opinion terms can be dually propagated from parameter learning in CRF to representation learning in RNN. We conduct expensive experiments on the dataset from SemEval challenge 2014 task 4 (subtask 1)~\cite{sem14} to verify the superiority of RNCRF over several baseline methods as well as the winning systems of the challenge.

\section{Related Work}

\subsection{Aspects and Opinions Co-Extraction}

Hu et al.~\shortcite{Hu04} proposed to extract product aspects through association mining, and opinion terms by augmenting a seed opinion set using synonyms and antonyms in WordNet. In follow-up work, syntactic relations are further exploited for aspect/opinion extraction~\cite{Pop05,Wu09,Qiu11}. For example, Qiu et al.~\shortcite{Qiu11} used syntactic relations to double propagate and augment the sets of aspects and opinions. Though the above models are unsupervised, 
they heavily depend on predefined rules for extraction, and are also restricted to specific types of POS tags for product aspects and opinions. Jin et al.~\shortcite{Jin09}, Li et al.~\shortcite{Li10}, Jakob et al.~\shortcite{Jakob10} and Ma et al.~\shortcite{Ma10} modeled the extraction problem as a sequence tagging problem, and proposed to use HMMs or CRFs to solve it. These methods rely on richly hand-crafted features, and do not consider interactions between aspect and opinion terms explicitly. Another direction is to use word alignment model to capture opinion relations among a sentence~\cite{Liu12,Liu13}. This method requires sufficient data for modeling desired relations.

Besides explicit aspects and opinions extraction, there are also other lines of research related to aspect-based sentiment analysis, including aspect classification~\cite{Hima14,Julian12}, aspect rating~\cite{Ivan08,wang11,Wang14}, domain-specific and target-dependent sentiment classification~\cite{Lu11,nir16,dong14,tang15}.

\subsection{Deep Learning for Sentiment Analysis}

Recent studies have shown that deep learning models can automatically learn the inherent semantic and syntactic information from data and thus achieve better performance for sentiment analysis~\cite{Socher11b,Socher12,Socher13,Glorot11,kalch14,kim14,Le14}. These methods generally belong to sentence-level or phrase/word-level sentiment polarity predictions. Regarding aspect-based sentiment analysis, Irsoy et al.~\shortcite{irsoy14} applied deep recurrent neural networks for opinion expression extraction. Dong et al.~\shortcite{dong14} proposed an adaptive recurrent neural network for target-dependent sentiment classification, where targets or aspects are given as input. Tang et al.~\shortcite{tang15} used Long Short-Term Memory (LSTM)~\cite{DBLP:journals/neco/HochreiterS97} for the same task. Nevertheless, there is little work in aspects and opinions co-extraction using deep learning models.

To the best of our knowledge, the most related works to ours are~\cite{liu15,Yin16}. Liu et al.~\shortcite{liu15} proposed to combine recurrent neural network and word embeddings to extract explicit aspects. However, the proposed model simply uses recurrent neural network on top of word embeddings, and thus its performance heavily depends on the quality of word embeddings. In addition, it fails to explicitly model dependency relations or compositionalities within certain syntactic structure in a sentence. Recently, Yin et al.~\shortcite{Yin16} proposed an unsupervised learning method to improve word embeddings using dependency path embeddings. A CRF is then trained with the embeddings independently in the pipeline.

Different from~\cite{Yin16}, our model does not focus on developing a new unsupervised word embedding methods, but encoding the information of dependency paths into RNN for constructing syntactically meaningful and discriminative hidden representations with labels. Moreover, we integrate RNN and CRF into a unified framework, and develop a joint optimization approach, instead of training word embeddings and a CRF separately as in~\cite{Yin16}. Note that Weiss et al.~\shortcite{weiss15} proposed to combine deep learning and structured learning for language parsing which can be learned by structured perceptron. However, they also separate neural network training with structured prediction.

Among deep learning methods, RNN has shown promising results on various NLP tasks, such as learning phrase representations~\cite{socher10}, sentence-level sentiment analysis~\cite{Socher13}, language parsing~\cite{Socher11a}, and question answering~\cite{DBLP:conf/emnlp/IyyerBCSD14}. The tree structures used for RNNs include constituency tree and dependency tree. In a constituency tree, all the words lie at leaf nodes, each internal node represents a phrase or a constituent of a sentence, and the root node represents the entire sentence~\cite{socher10,Socher12,Socher13}. In a dependency tree, each node including terminal and nonterminal nodes, represents a word, with dependency connections to other nodes~\cite{Socher14,DBLP:conf/emnlp/IyyerBCSD14}. The resultant model is known as dependency-tree RNN (DT-RNN). An advantage of using dependency tree over the other is the ability to extract word-level representations considering syntactic relations and semantic robustness. Therefore, we adopt DT-RNN in this work.

\section{Problem Statement}\label{sec:definition}
Suppose that we are given a training set of customer reviews in a specific domain, denoted by $S\!=\!\{s_{1}, ... , s_{N}\}$, where $N$ is the number of review sentences. For any $s_{i}\!\in\! S$, there may exist a set of aspect terms $A_{i}\!=\!\{a_{i1}, ... , a_{il}\}$, where each $a_{ij}\!\in\! A_i$ can be a single word or a sequence of words expressing explicitly some aspect of an entity, and a set of opinion terms $O_{i}\!=\!\{o_{i1}, ... , o_{im}\}$, where each $o_{ir}$ can be a single word or a sequence of words expressing the subjective sentiment of the comment holder. The task is to learn a classifier to extract the set of aspect terms $A_{i}$ and the set of opinion terms $O_{i}$ from each review sentence $s_{i}\!\in\! S$.

This task can be formulated as a sequence tagging problem by using the BIO encoding scheme. Specifically, each review sentence $s_{i}$ is composed of a sequence of words $s_{i}\!=\!\{w_{i1}, ... , w_{in_{i}}\}$. Each word $w_{ip}\!\in\! s_{i}$ is labeled as one out of the following 5 classes: ``BA'' (beginning of aspect), ``IA'' (inside of aspect), ``BO'' (beginning of opinion), ``IO'' (inside of opinion), and ``O'' (others). Let ${\bf L}\!=\!\{\mbox{BA},\mbox{IA},\mbox{BO},\mbox{IO},\mbox{O}\}$. We are also given a test set of review sentences denoted by $S'\!=\!\{s'_{1}, ... , s'_{N'}\}$, where $N'$ is the number of test reviews. For each test review $s'_{i}\!\in\! S'$, our objective is to predict the class label $y'_{iq}\!\in\!{\bf L}$ for each word $w'_{iq}$. Note that a sequence of predictions with ``BA'' at the beginning followed by ``IA'' are indication of one aspect, which is similar for opinion terms.\footnote{In this work we focus on extraction of aspect and opinion terms, not polarity predictions on opinion terms. Polarity prediction can be done by either post-processing on the extracted opinion terms or redefining the BIO labels by encoding the polarity information.}

\begin{figure*}[t!]
	\centering
	\subfigure[Example of a dependency tree.]{\label{fig:DT}\includegraphics[width=0.315\textwidth]{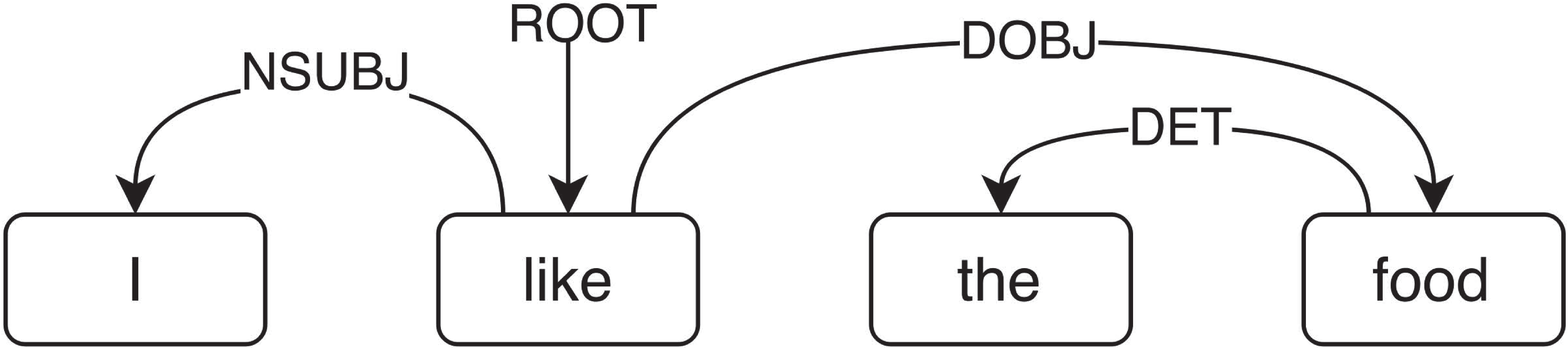}}\hspace{3mm}
	\subfigure[Example of a DT-RNN tree structure.]{\label{fig:RNN}\includegraphics[width=0.315\textwidth]{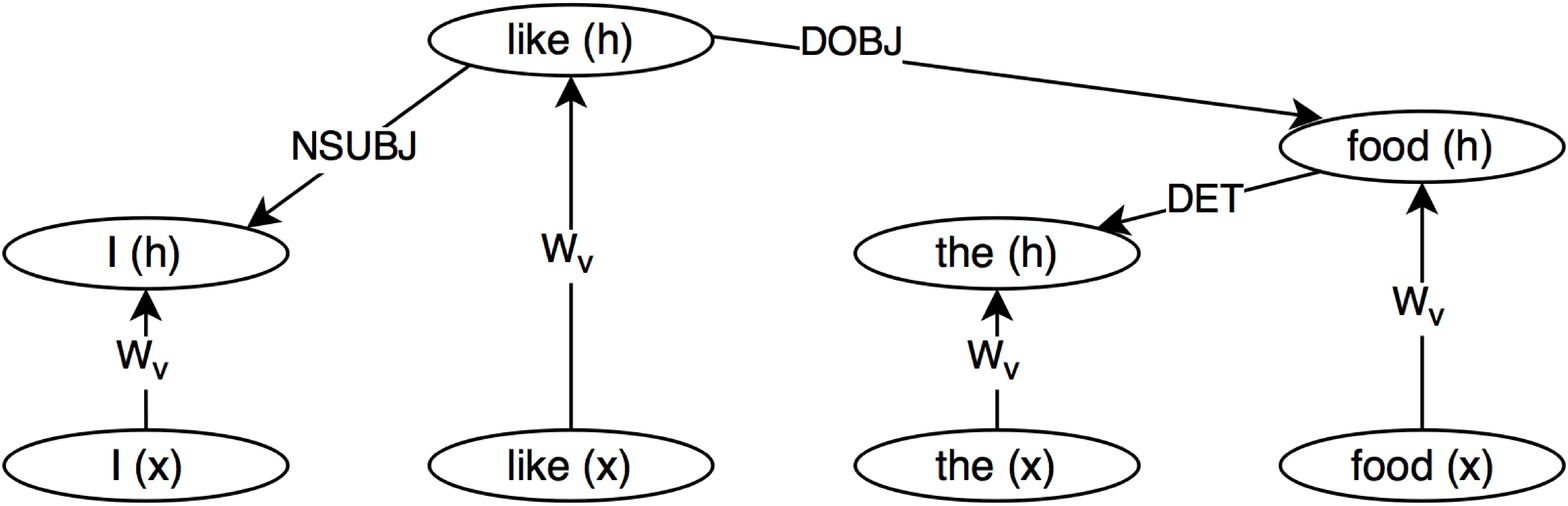}}\hspace{3mm}
	\subfigure[Example of a RNCRF structure.]{\label{fig:RNCRF}\includegraphics[width=0.315\textwidth]{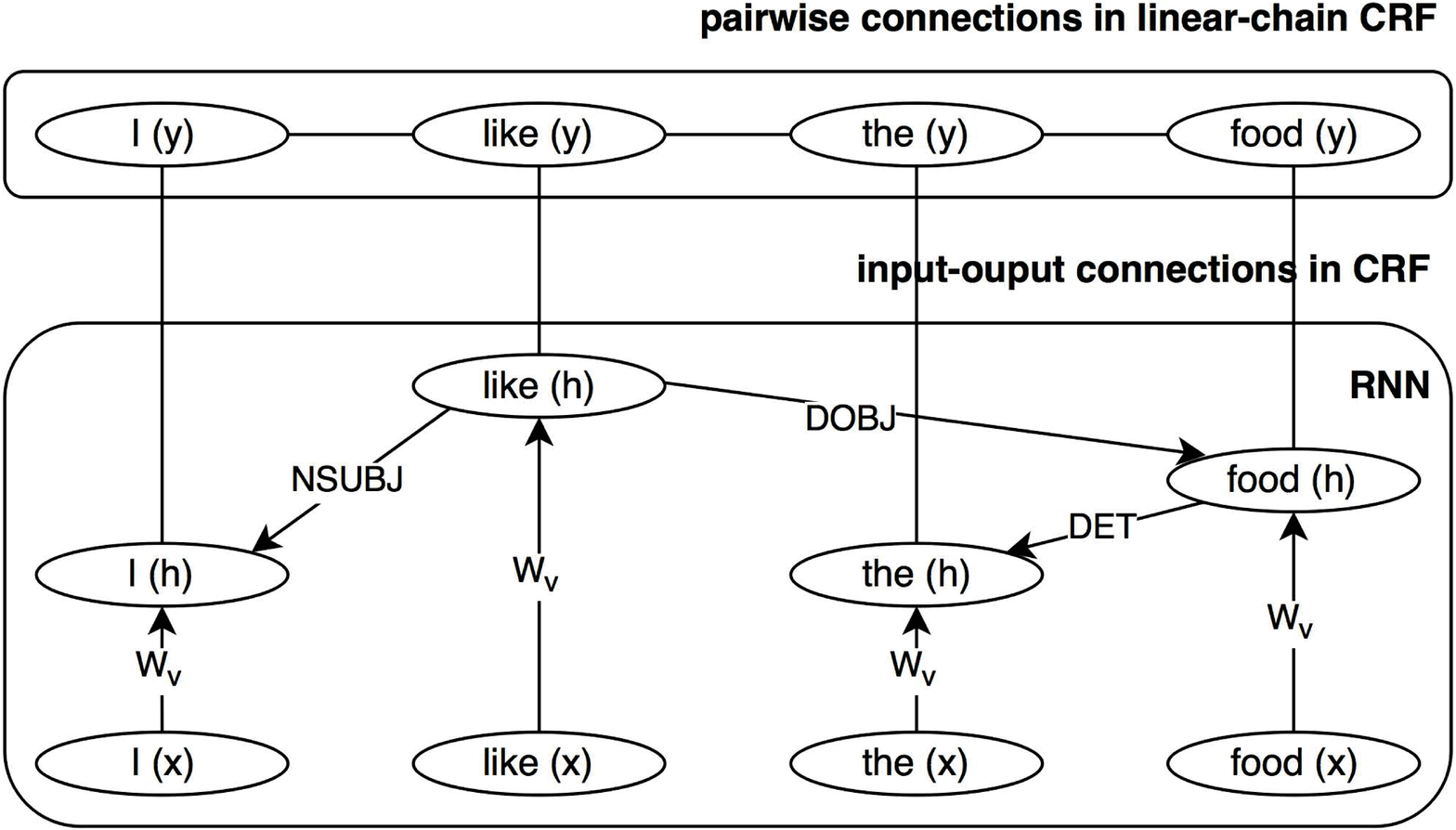}}
	\caption{Examples of dependency tree, DT-RNN structure and RNCRF structure for a review sentence.}\label{fig:model:overview}
\end{figure*}

\section{Recursive Neural CRFs}

As described in Section~\ref{sec:intro}, RNCRF consists of two main components: 1) a DT-RNN to learn a high-level representation for each word in a sentence, and 2) a CRF to take the learned representation as input to capture context around each word for explicit aspect and opinion terms extraction. Next, We present these two components in details.

\subsection{Dependency-Tree RNNs}
We begin by associating each word $w$ in our vocabulary with a feature vector $x\!\in\!\R^{d}$, which corresponds to a column of a word embedding matrix $W_e\!\in\!\R^{d\times v}$, where $v$ is the size of the vocabulary. For each sentence, we build a DT-RNN based on the corresponding dependency parse tree with word embeddings as initialization. An example of the dependency parse tree is shown in Figure~\ref{fig:DT}, where each edge starts from the parent and points to its dependent with a syntactic relation. 

In a DT-RNN, each node $n$, including leaf nodes, internal nodes and the root node, in a specific sentence is associated with a word $w$, an input feature vector $x_w$, and a hidden vector $h_n\!\in\!\R^{d}$ of the same dimension as $x_w$. Each dependency relation $r$ is associated with a separate matrix $W_r\!\in\!\R^{d\times d}$. In addition, a common transformation matrix $W_v\!\in\!\R^{d\times d}$ is introduced to map the word embedding $x_w$ at node $n$ to its corresponding hidden vector $h_n$.

Along with a particular dependency tree, a hidden vector $h_n$ is computed from its own word embedding $x_w$ at node $n$ with the transformation matrix $W_v$ and its children's hidden vectors $h_{\textup{child(n)}}$ with the corresponding relation matrices $\{W_r\}$'s. For instance, given the parse tree shown in Figure~\ref{fig:DT}, we first compute the leaf nodes associated with the words \textit{I} and \textit{the} using $W_v$ as follows,
\begin{eqnarray}
h_{\textup{I}} & = & f(W_{v}\cdot x_{\textup{I}} + b), \nonumber \\
h_{\textup{the}} & = & f(W_{v}\cdot x_{\textup{the}} + b), \nonumber
\end{eqnarray}
where $f$ is a nonlinear activation function and $b$ is a bias term. In this paper, we adopt $tanh(\cdot)$ as the activation function. Once the hidden vectors of all the leaf nodes are generated, we can recursively generate hidden vectors for interior nodes using the corresponding relation matrix $W_r$ and the common transformation matrix $W_v$ as follows,
\begin{eqnarray}
h_{\textup{food}} & = & f(W_{v}\cdot x_{\textup{food}} + W_{\textup{DET}}\cdot h_{\textup{the}} + b), \nonumber \\
h_{\textup{like}} & = & f(W_{v}\cdot x_{\textup{like}} + W_{\textup{DOBJ}}\cdot h_{\textup{food}} \nonumber \\
& &\;\;\; + W_{\textup{NSUBJ}} \cdot h_{\textup{I}} + b). \nonumber
\end{eqnarray}
The resultant DT-RNN is shown in Figure~\ref{fig:RNN}. In general, a hidden vector for any node $n$ associated with a word vector $x_w$ can be computed as follows,
\begin{equation}\label{eq:dt_rnn}
h_n = f\left(W_v\cdot x_w + b + \sum_{k\in\K_n}W_{r_{nk}}\cdot h_k\right),
\end{equation}
where $\K_n$ denotes the set of children of node $n$, $r_{nk}$ denotes the dependency relation between node $n$ and its child node $k$, and $h_k$ is the hidden vector of the child node $k$. The parameters of DT-RNN, $\Theta_{\textup{RNN}} \!=\! \{W_{v}, W_{r}, W_{e}, b\}$, are learned during training.

\subsection{Integration with CRFs}\label{sec:crf}

CRFs are a discriminant graphical model for structured prediction. In RNCRF, we feed the output of DT-RNN, i.e., the hidden representation of each word in a sentence, to a CRF. Updates of parameters for RNCRF are carried out successively from the top to bottom, by propagating errors through CRF to the hidden layers of RNN (including word embeddings) using backpropagation through structure (BPTS)~\cite{Goller96}.

Formally, for each sentence $s_i$, we denote the input for CRF by ${\bf h}_{i}$, which is generated by DT-RNN. Here ${\bf h}_{i}$ is a matrix with columns of hidden vectors $\{h_{i1}, ..., h_{in_{i}}\}$ to represent a sequence of words $\{w_{i1},...,w_{in_i}\}$ in a sentence $s_i$. The model computes a structured output ${\bf y}_{i}=\{y_{i1}, ... , y_{in_{i}}\}\!\in\! {\bf Y}$, where ${\bf Y}$ is a set of possible combinations of labels in label set ${\bf L}$. The entire structure can be represented by an undirected graph $G=(V,E)$ with cliques $c\!\in\! C$. In this paper, we employed linear-chain CRF, which has two different cliques: unary clique (U) representing input-output connection, and pairwise clique (P) representing adjacent output connection, as shown in Figure~\ref{fig:RNCRF}. During inference, the model aims to output ${\bf \hat{y}}$ with the maximum conditional probability $p({\bf y}|{\bf h})$. (We drop the subscript $i$ here for simplicity.) The probability is computed from potential outputs of the cliques:
\begin{equation}\label{eq:crf}
p({\bf y}|{\bf h})=\frac{1}{Z({\bf h})}\prod_{c\in C}\psi_{c}({\bf h},{\bf y}_{c}),
\end{equation}
where $Z({\bf h})$ is the normalization term, and $\psi_{c}({\bf h},{\bf y}_{c})$ is the potential of clique $c$, computed as
$\psi_{c}({\bf h},{\bf y}_{c}) \!=\! \exp\left \langle W_{c},F({\bf h},{\bf y}_{c}) \right \rangle$,
where the RHS is the exponential of a linear combination of feature vector $F({\bf h},{\bf y}_{c})$ for clique $c$, and the weight vector $W_{c}$ is tied for unary and pairwise cliques. We also incorporate a context window of size $2T\!+\!1$ when computing unary potentials. Thus, the potential of unary clique at node $k$ can be written as
\begin{eqnarray}
\psi_{U}({\bf h},y_{k}) = \exp \!\!\!&\!\!\! \!\!\!&\!\!\! \left((W_{0})_{y_{k}} \!\cdot\! h_{k} + \sum_{t=1}^T(W_{-t})_{y_{k}} \!\cdot\! h_{k-t}\right. \nonumber \\
\!\!\!&\!\!\! \!\!\!&\!\!\! \left.+ \sum_{t=1}^T(W_{+t})_{y_{k}}\cdot h_{k+t}\right),
\end{eqnarray}
where $W_{0}$, $W_{+t}$ and $W_{-t}$ are weight matrices of the CRF for the current position, the $t$-th position to the right, and the $t$-th position to the left within context window, respectively. The subscript $y_{k}$ indicates the corresponding row in the weight matrix.

\begin{figure}
	\centering
	\includegraphics[width=0.825\linewidth]{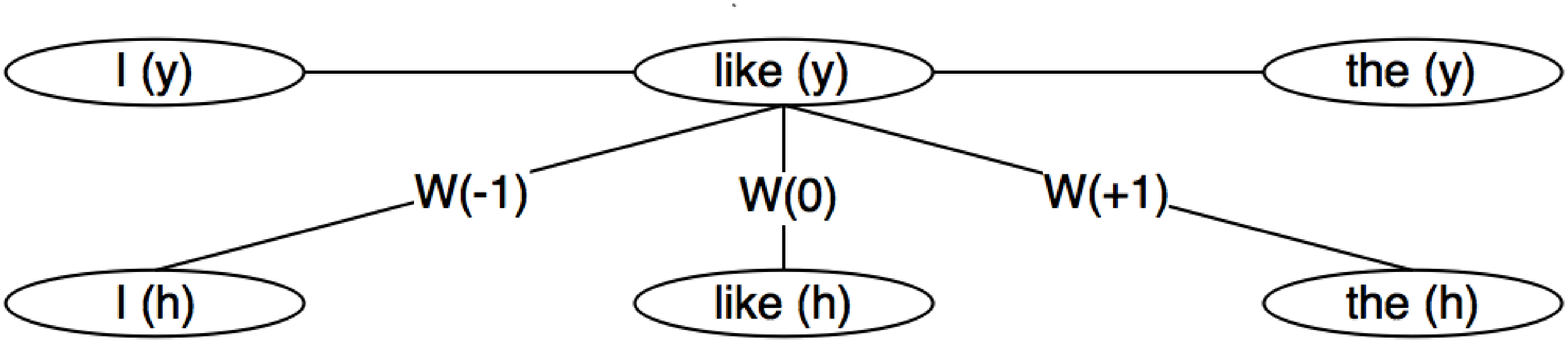}
	\caption{An example for computing input-ouput potential for the second position \textit{like}.}
	\label{fig:detail_rnncrf_mod2}
\end{figure}
For instance, Figure~\ref{fig:detail_rnncrf_mod2} shows an example of window size 3. At the second position, the input features for \textit{like} are composed of the hidden vectors at position 1 ($h_{\textup{I}}$), position 2 ($h_{\textup{like}}$) and position 3 ($h_{\textup{the}}$). Therefore, the conditional distribution for the entire sequence ${\bf y}$ in Figure~\ref{fig:RNCRF} can be calculated as
\begin{small}
\begin{eqnarray}
p({\bf y}|{\bf h}) \!\!=\!\! \frac{1}{Z({\bf h})}\exp \!\!\!\!\!\!&\!\!\!\!\!\! \!\!&\!\!
\left(\sum_{k=1}^4(W_{0})_{y_{k}}\!\cdot\! h_{k}
\!\!+\!\!\sum_{k=2}^4(W_{-1})_{y_{k}}\!\cdot\! h_{k-1}\right. 
\nonumber \\
\!\!&\!\! \!\!&\!\!\left.\!\!+\!\! \sum_{k=1}^{3}(W_{+1})_{y_{k}}\!\cdot\! h_{k+1}\!\!+\!\!\sum_{k=1}^{3}V_{y_{k},y_{k+1}}\right), \nonumber
\end{eqnarray}
\end{small}
$\!\!$where the first three terms in the exponential of the RHS consider unary clique while the last term considers the pairwise clique with matrix $V$ representing pairwise state transition score. For simplicity in description on parameter updates, we denote the log-potential for clique $c\!\in\!\{U,P\}$ by $g_{c}({\bf h},{\bf y}_{c})\!=\!\left \langle W_{c},F({\bf h},{\bf y}_{c}) \right\rangle$.

\subsection{Joint Training for RNCRF}\label{sec:training}
Through the objective of maximum likelihood, updates for parameters of RNCRF are first conducted on the parameters of the CRF (unary weight matrices $\Theta_{U}=\{W_{0}, W_{+t}, W_{-t}\}$ and pairwise weight matrix $V$) by applying chain rule to log-potential updates. Below is the gradient for $\Theta_{U}$ (updates for $V$ are similar through the log-potential of pairwise clique $g_{P}(y_{k}',y_{k+1}')$):
\begin{equation}\label{eq:paramUpdate:1}
\triangle\Theta_{U} = \frac{\partial -\log p({\bf y}|{\bf h})}{\partial g_{U}({\bf h},y_{k}')}\!\cdot\!\frac{\partial g_{U}({\bf h},y_{k}')}{\partial\Theta_{U}},
\end{equation}
where
\begin{equation}\label{eq:paramUpdate:2}
\frac{\partial -\log p({\bf y}|{\bf h})}{\partial g_{U}({\bf h},y_{k}')} = -(1_{y_{k}=y_{k}'}-p(y_{k}'|{\bf h})),
\end{equation}
and $y_{k}'$ represents possible label configuration of node $k$. The hidden representations of each word and the parameters of DT-RNN are updated subsequently by applying chain rule with (\ref{eq:paramUpdate:2}) through BPTS as follows,
\begin{eqnarray}
\!\!\!\!\!\!\triangle h_{\textup{root}} \!&\!\! = \!\!&\! \frac{\partial -\log p({\bf y}|{\bf h})}{\partial g_{U}({\bf h},y_{\textup{root}}')}\cdot\frac{\partial g_{U}({\bf h},y_{\textup{root}}')}{\partial h_{\textup{root}}}, \label{eq:paramUpdate:3} \\
\!\!\!\!\!\!\triangle h_{k\neq\textup{root}} \!&\!\! = \!\!&\! \frac{\partial -\log p({\bf y}|{\bf h})}{\partial g_{U}({\bf h},y_{k}')}\cdot\frac{\partial g_{U}({\bf h},y_{k}')}{\partial h_{k}} \nonumber\\
\!\!\!\!\!\!&\! \!& + \triangle h_{\textup{\textup{par}(k)}}\cdot\frac{\partial h_{\textup{par}(k)}}{\partial h_{k}}, \label{eq:paramUpdate:4} \\
\!\!\!\!\!\!\triangle\Theta_{\textup{RNN}} \!&\!\! = \!\!&\! \sum_{k=1}^K\frac{\partial -\log p({\bf y}|{\bf h})}{\partial h_{k}}\cdot\frac{\partial h_{k}}{\partial\Theta_{\textup{RNN}}}, \label{eq:paramUpdate:5}
\end{eqnarray}
where $h_{\textup{root}}$ represents the hidden vector of the word pointed by ROOT in the corresponding DT-RNN. Since this word is the topmost node in the tree, it only inherits error from the CRF output. In (\ref{eq:paramUpdate:4}), $h_{\textup{par}(k)}$ denotes the hidden vector of the parent node of node $k$ in DT-RNN. Hence the lower nodes receive error from both the CRF output and error propagation from parent node. The parameters within DT-RNN, $\Theta_{\textup{RNN}}$, are updated by applying chain rule with respect to updates of hidden vectors, and aggregating among all associated nodes, as shown in (\ref{eq:paramUpdate:5}). The overall procedure of RNCRF is summarized in Algorithm~\ref{alg:rncrf}.

\begin{algorithm}[t!]
\footnotesize
\caption{Recursive Neural CRFs}\label{alg:rncrf}
	\begin{algorithmic}
		\STATE {\bf Input:} A set of customer review sequences: $S=\{s_{1}, ... , s_{N}\}$, and feature vectors of $d$ dimensions for each word $\{x_w\}$'s, window size $T$ for CRFs
		\STATE {\bf Output:} Parameters: $\Theta \!=\! \left\{\Theta_{\textup{RNN}}, \Theta_{U}, V\right\}$
		\STATE {\bf Initialization:} Initialize $W_{e}$ using {\em word2vec}. Initialize $W_v$ and $\{W_r\}$'s randomly with uniform distribution between $\left[-{\frac{\sqrt{6}}{\sqrt{2d+1}}},\frac{\sqrt{6}}{\sqrt{2d+1}}\right]$. Initialize $W_0$, $\{W_{+t}\}$'s, $\{W_{-t}\}$'s, $V$, and $b$ with all 0's
		\FOR {each sentence $s_i$}
		\STATE 1: Use DT-RNN (\ref{eq:dt_rnn}) to generate ${\bf h}_i$
		\STATE 2: Compute $p({\bf y}_i|{\bf h}_i)$ using (\ref{eq:crf})
		\STATE 3: Use the backpropagation algorithm to update parameters $\Theta$ through (\ref{eq:paramUpdate:1})-(\ref{eq:paramUpdate:5})
		\ENDFOR
	\end{algorithmic}
\end{algorithm}

\section{Discussion}
The best performing system~\cite{toh14} for SemEval challenge 2014 task 4 (subtask 1) employed CRFs with extensive hand-crafted features including those induced from dependency trees. However, their experiments showed that the addition of the features induced from dependency relations does not improve the performance. This indicates the infeasibility or difficulty of incorporating dependency structure explicitly as input features, which motivates the design of our model to use DT-RNN to encode dependency between words for feature learning. The most important advantage of RNCRF is the ability to learn the underlying dual propagation between aspect and opinion terms from the tree structure itself. Specifically as shown in Figure~\ref{fig:RNCRF}, where the aspect is \textit{food} and the opinion expression is \textit{like}. In the dependency tree, \textit{food} depends on \textit{like} with the relation \textit{DOBJ}. During training, RNCRF computes the hidden vector $h_{\textup{like}}$ for \textit{like}, which is also obtained from $h_{\textup{food}}$. As a result, the prediction for \textit{like} is affected by $h_{\textup{food}}$. This is one-way propagation from \textit{food} to \textit{like}. During backpropagation, the error for \textit{like} is propagated through a top-down manner to revise the representation $h_{\textup{food}}$. This is the other-way propagation from \textit{like} to \textit{food}. Therefore, the dependency structure together with the learning approach help to enforce the dual propagation of aspect-opinion pairs as long as the dependency relation exists, either directly or indirectly.

\subsection{Adding Linguistic/Lexicon Features}\label{sec:feature}
RNCRF is an end-to-end model, where feature engineering is not necessary. However, it is flexible to incorporate \textit{light} hand-crafted features into RNCRF to further boost its performance, such as features with POS tags, name-list, or sentiment lexicon. These features could be appended to the hidden vector of each word, but keep fixed during training, unlike learnable neural inputs and the CRF weights as described in Section~\ref{sec:training}. As will be shown in experiments, RNCRF without any hand-crafted features slightly outperforms the best performing systems that involve heavy feature engineering efforts, and RNCRF with light feature engineering can achieve better performance.

\section{Experiment}

\subsection{Dataset and Experimental Setup}
We evaluate our model on the dataset from SemEval Challenge 2014 task 4 (subtask 1), which includes reviews from two domains: restaurant and laptop reviews\footnote{Experiments with more publicly available datasets, e.g. restaurant review dataset from SemEval Challenge 2015 task 12 will be conducted in our future work.}. The detailed description of the dataset is given in Table~\ref{tbl:dataset}. As the original dataset only includes manually annotate labels for aspect terms but not for opinion terms, we manually annotated opinion terms for each sentence by ourselves to facilitate our experiments.

For word vector initialization, we train word embeddings with word2vec~\cite{DBLP:journals/corr/abs-1301-3781} on the Yelp Challenge dataset\footnote{http://www.yelp.com/dataset\_challenge} for the restaurant domain and on the Amazon reviews\footnote{http://jmcauley.ucsd.edu/data/amazon/links.html}~\cite{Mc15} for the laptop domain. The Yelp dataset contains 2.2M restaurant reviews with 54K vocabulary size. For the Amazon reviews, we only extracted the electronic domain that contains 1M reviews with 590K vocabulary size. We vary different dimensions for word embeddings and chose 300 for both domains. Empirical sensitivity studies on different dimensions of word embeddings are also conducted.
\begin{table}[t]
\footnotesize
	\begin{center}
		\begin{tabular}{l|c|c|c}
			\hline
			\hline Domain     & Training  & Test  & Total \\
			\hline Restaurant & 3,041     & 800   & 3,841 \\
			\hline Laptop     & 3,045     & 800   & 3,845 \\
			\hline Total      & 6,086     & 1,600 & 7,686 \\
			\hline
		\end{tabular}
		\caption{SemEval Challenge 2014 task 4 dataset}\label{tbl:dataset}
	\end{center}
\end{table}
Dependency trees are generated using Stanford Dependency Parser~\cite{Klein03}. Regarding CRFs, we implement a linear-chain CRF using CRFSuite~\cite{CRFsuite}. Because of the relatively small size of training data and a large number of parameters, we perform pretraining on the parameters of DT-RNN with cross-entropy error, which is a common strategy for deep learning~\cite{Erhan09}. We implement mini-batch stochastic gradient descent (SGD) with a batch size of 25, and an adaptive learning rate (AdaGrad) initialized at 0.02 for pretraining of DT-RNN, which runs 4 epochs for the restaurant domain and 5 epochs for the laptop domain. For parameter learning of the joint model RNCRF, we implement SGD with a decaying learning rate initialized at 0.02. We also try with varying context window size, and use 3 for the laptop domain and 5 for the restaurant domain, respectively. All parameters are chosen by cross validation.

As discussed in Section~\ref{sec:feature}, hand-crafted features can be easily incorporated into RNCRF. We generate three types of simple features based on POS tags, name-list and sentiment lexicon to show further improvement by incorporating these features. Following~\cite{toh14}, we extract two sets of name list from the training data for each domain, where one includes high-frequency aspect terms, and the other includes high-probability aspect words. These two sets are used to construct two lexicon features, i.e. we build a 2D binary vector: if a word is in a set, the corresponding value is 1, otherwise 0. For POS tags, we use Stanford POS tagger~\cite{Tout03}, and convert them to universal POS tags that have 15 different categories. We then generate 15 one-hot POS tag features. For sentiment lexicon, we use the collection of commonly used opinion words (around 6,800)~\cite{Hu04}. Similar to name list, we create a binary feature to indicate whether the word belongs to opinion lexicon. We denote by RNCRF+F the proposed model with the three types of features.

Compared to the winning systems of SemEval Challenge 2014 task 4 (subtask 1), RNCRF or RNCRF+F uses additional labels of opinion terms for training. Therefore, to conduct fair comparison experiments with the winning systems, we implement RNCRF-O by omitting opinion labels to train our model (i.e., labels become ``BA'', ``IA'', ``O''). Accordingly, we denote by RNCRF-O+F the RNCRF-O model with the three additional types of hand-crafted features.

\subsection{Experimental Results}
We compare our model with several baselines:
\begin{itemize}
\item \textbf{CRF-1}: a linear-chain CRF with standard linguistic features including word string, stylistics, POS tag, context string, and context POS tags.
\item \textbf{CRF-2}: a linear-chain CRF with both standard linguistic features and dependency information including head word, dependency relations with parent token and child tokens.
\item \textbf{LSTM}: an LSTM network built on top of word embeddings proposed by~\cite{liu15}. We keep original settings in~\cite{liu15} but replace their word embeddings with ours (300 dimension).
We try different hidden layer dimensions (50, 100, 150, 200) and reported the best result with size 50.
\item \textbf{LSTM+F}: the above LSTM model with the three additional types of hand-crafted features as with RNCRF.
\item \textbf{SemEval-1}, \textbf{SemEval-2}: the top two winning systems for SemEval challenge 2014 task 4 (subtask 1).
\item \textbf{WDEmb+B+CRF}\footnote{We report the best results from the original paper~\cite{Yin16}.}: the model proposed by~\cite{Yin16} using word and dependency path embeddings combined with linear context embedding features, dependency context embedding features and hand-crafted features (i.e., feature engineering) as CRF input.
\end{itemize}
The comparison results are shown in Table~\ref{tbl:comparison} for both the restaurant domain and the laptop domain. Note that we provide the same annotated dataset (both aspect labels and opinion labels are included for training) for CRF-1, CRF-2 and LSTM for fair comparison. It is clear that our proposed model RNCRF achieves superior performance compared with most of the baseline models. The performance is even better by adding simple hand-crafted features, i.e., RNCRF+F, with 0.92\% and 3.87\% absolute improvement over the best system in the challenge for aspect extraction for the restaurant domain and the laptop domain, respectively. This shows the advantage of combining high-level continuous features and discrete hand-crafted features. Though CRFs usually show promising results in sequence tagging problems, they fail to achieve comparable performance when lacking of extensive features (e.g., CRF-1). By adding dependency information explicitly in CRF-2, the result only improves slightly for aspect extraction. Alternatively, by incorporating dependency information into a deep learning model (e.g., RNCRF), the result shows more than 7\% improvement for aspect extraction and 2\% for opinion extraction.

\begin{table}
\footnotesize
	\begin{center}
		\begin{tabular}{p{2.2cm}|c|c|c|p{1cm}}
			\hline
			\hline & \multicolumn{2}{c|}{Restaurant} & \multicolumn{2}{c}{Laptop} \\
			\hline Models & Aspect & Opinion & Aspect & Opinion \\
            \hline SemEval-1 & 84.01 & - & 74.55 & - \\
			\hline SemEval-2 & 83.98 & - & 73.78 & - \\
            \hline WDEmb+B+CRF & \bf{84.97} & - & 75.16 & - \\
			\hline CRF-1 & 77.00 & 78.95 & 66.21 & 71.78 \\
			\hline CRF-2 & 78.37 & 78.65 & 68.35 & 70.05 \\
			\hline LSTM & 81.15 & 80.22 & 72.73 & 74.98 \\
            \hline LSTM+F & 82.99 & 82.90 & 73.23 & 77.67 \\
            \hline
			\hline RNCRF-O & 82.73 & - & 74.52 & - \\
			\hline RNCRF-O+F & 84.25 & - & 77.26 & - \\
			\hline RNCRF & 84.05 & 80.93 & 76.83 & 76.76 \\
			\hline
			RNCRF+F & 84.93 & \bf{84.11} & \bf{78.42} & \bf{79.44} \\
			\hline
		\end{tabular}
		\caption{Comparison results in terms of F1 scores.}\label{tbl:comparison}
	\end{center}
\end{table}

By removing the labels for opinion terms, RNCRF-O produces inferior results than RNCRF because the effect of dual propagation of aspect and opinion pairs disappears with the absence of opinion labels. This verifies our previous assumption that DT-RNN could learn the interactive effects within aspects and opinions. However, the performance of RNCRF-O is still comparable to the top systems and even better with the addition of simple linguistic features: 0.24\% and 2.71\% superior than the best system in the challenge for the restaurant domain and the laptop domain, respectively. This shows the robustness of our model even without additional opinion labels.

LSTM has shown comparable results for aspect extraction~\cite{liu15}. However, in their work, they used well-pretrained word embeddings by training with large corpus or extensive external resources, e.g. chunking, and NER. To compare their model with RNCRF, we re-implement LSTM with the same word embedding strategy and labeling resources as ours. The results show that our model outperforms LSTM in aspect extraction by 2.90\% and 4.10\% for the restaurant domain and the laptop domain, respectively. We conclude that a standard LSTM model fails to extract the relations between aspect and opinion terms. Even with the addition of same linguistic features, LSTM is still inferior than RNCRF itself in terms of aspect extraction. Moreover, our result is comparable with WDEmb+B+CRF in the restaurant domain and better in the laptop domain (+3.26\%). Note that WDEmb+B+CRF appended dependency context information into CRF while our model encode such information into high-level representation learning.

\begin{table}[t]
\footnotesize
	\begin{center}
		\begin{tabular}{l|c|c|c|p{1cm}}
			\hline
			\hline & \multicolumn{2}{c|}{Restaurant} & \multicolumn{2}{c}{Laptop} \\
			\hline Models & $\!$Aspect$\!$ & $\!$Opinion$\!$ & $\!$Aspect$\!$ & $\!$Opinion$\!$ \\
			\hline DT-RNN $\!$+$\!$ SoftMax$\!\!$ &
            \multirow{1}{*}{72.45} &
            \multirow{1}{*}{69.76} &
            \multirow{1}{*}{66.11} &
            \multirow{1}{*}{64.66} \\
			\hline CRF $\!$+$\!$ word2vec &
            \multirow{1}{*}{82.57} &
            \multirow{1}{*}{78.83} &
            \multirow{1}{*}{63.62} &
            \multirow{1}{*}{56.96} \\
			\hline RNCRF & 84.05 & 80.93 & 76.83 & 76.76 \\
			\hline RNCRF+POS & 84.08 & 81.48 & 77.04 & 77.45\\
			\hline RNCRF+NL & 84.24 & 81.22 & 78.12 & 77.20 \\
			\hline RNCRF+Lex & 84.21 & \bf{84.14} & 77.15 & 78.56\\
            \hline RNCRF+F & \bf{84.93} & 84.11 & \bf{78.42} & \bf{79.44} \\
			\hline
		\end{tabular}
		\caption{Impact of different components.}\label{tbl:RNCRF:components}
	\end{center}
\end{table}

To test the impact of each component of RNCRF and the three types of hand-crafted features, we conduct experiments on different model settings:
\begin{itemize}
\item \textbf{DT-RNN+SoftMax}: rather than using a CRF, a softmax classifier is used on top of DT-RNN.
\item \textbf{CRF+word2vec}: a linear-chain CRF with word embeddings only without using DT-RNN.
\item \textbf{RNCRF+POS/NL/Lex}: the RNCRF model with POS tag or name list or sentiment lexicon feature(s).
\end{itemize}
The comparison results are shown in Table~\ref{tbl:RNCRF:components}. Similarly, both aspect and opinion term labels are provided for training for each of the above models. Firstly, RNCRF achieves much better results compared to DT-RNN+SoftMax (+11.60\% and +10.72\% for the restaurant domain and the laptop domain in aspect extraction). This is because DT-RNN fails to fully exploit context information for sequence labeling, which, in contrast, can be achieved by CRF. Secondly, RNCRF outperforms CRF+word2vec, which proves the importance of DT-RNN for modeling interactions between aspects and opinions. Hence, the combination of DT-RNN and CRF inherits the advantages from both models. Moreover, by separately adding hand-crafted features, we can observe that name-list based features and the sentiment lexicon feature are most effective for aspect extraction and opinion extraction, respectively. This may be explained by the fact that name-list based features usually contain informative evident for aspect terms and sentiment lexicon provides explicit indication about opinions. 

\begin{figure}[t!]
\centering
	\subfigure[On the restaurant domain.]{\label{fig:res_sensitivity_new}\includegraphics[width=0.755\columnwidth]{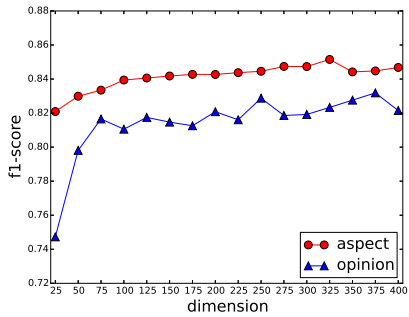}}\\
	\subfigure[On the laptop domain.]{\label{fig:lap_sensitivity_new}\includegraphics[width=0.755\columnwidth]{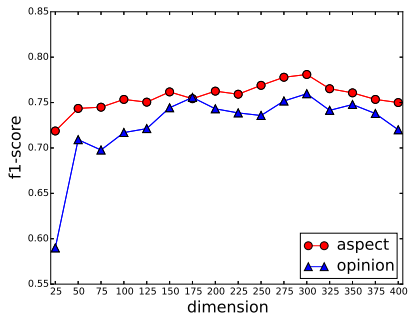}}
	\caption{Sensitivity studies on word embeddings.}\label{fig:sensitivity_new}
\end{figure}
Besides the comparison experiments, we also conduct sensitivity test for our proposed model in terms of word vector dimensions. We tested a set of different dimensions ranging from 25 to 400, with 25 increment. The sensitivity plot is shown in Figure~\ref{fig:sensitivity_new}. The performance for aspect extraction is smooth with different vector lengths for both domains. For restaurant domain, the result is stable when dimension is larger than or equal to 100, with the highest at 325. For the laptop domain, the best result is at dimension 300, but with relatively small variations. For opinion extraction, the performance reaches a good level when the dimension is larger than or equal to 75 for the restaurant domain and 125 for the laptop domain. This proves the stability and robustness of our model.

\section{Conclusion}
We have presented a joint model, RNCRF, that achieves the state-of-the-art performance for explicit aspect and opinion term extraction on a benchmark dataset. With the help of DT-RNN, high-level features can be learned by encoding the underlying dual propagation of aspect-opinion pairs. 
RNCRF combines the advantages of DT-RNNs and CRFs, and thus outperforms the traditional rule-based methods in terms of flexibility, because aspect terms and opinion terms are not only restricted to certain observed relations and POS tags. Compared to feature engineering methods with CRFs, the proposed model saves much effort in composing features, and it is able to extract higher-level features obtained from non-linear transformations.

\section*{Acknowledgements}
This research is partially funded by the Economic Development Board and the National Research Foundation of Singapore. Sinno J. Pan thanks the support from Fuji Xerox Corporation through joint research on {\em Multilingual Semantic Analysis} and the NTU Singapore Nanyang Assistant Professorship (NAP) grant M4081532.020.

\bibliography{emnlp2016}
\bibliographystyle{emnlp2016}

\end{document}